\documentclass[10pt,twocolumn,letterpaper]{article}

\usepackage{iccv}
\usepackage{times}
\usepackage{epsfig}
\usepackage{graphicx}
\graphicspath{{./}{./Figures/}}
\usepackage{amsmath}
\usepackage{amssymb}
\usepackage{subcaption}
\usepackage{booktabs}

\newif\ifrevfinal
\revfinalfalse
\def\rev[#1][#2]{\ifrevfinal #2 \else {\color{blue} \sout{#1}} {\bf \color{red} #2} \fi}
\usepackage{amsmath,amssymb,amsbsy,xspace}

\def\model#1#2{{#1}{\scriptsize2}{#2}}

\usepackage{amsmath}

\def\<{\langle}
\def\>{\rangle}

\def\be{\begin{equation}}
\def\ee{\end{equation}}
\def\bea{\begin{eqnarray}}
\def\eea{\end{eqnarray}}
\def\tab#1{Table~\ref{tab:#1}}
\def\sect#1{Section~\ref{sec:#1}}

\makeatletter
\gdef\SetFigFont#1#2#3{%
  \reset@font\fontsize{10}{12pt}%
  \selectfont%
}
\let\eqnarray@=\eqnarray \let\endeqnarray@=\endeqnarray
\def\eqnarray{\bgroup\arraycolsep=2pt\eqnarray@}
\def\endeqnarray{\endeqnarray@\egroup}
\makeatother
\def\arraystretch{1.2}
\def\em{\slshape}

\makeatletter
\DeclareRobustCommand\onedot{\futurelet\@let@token\@onedot}
\def\@onedot{\ifx\@let@token.\else.\null\fi\xspace}

\def\eg{\emph{e.g}\onedot} 
\def\ie{\emph{i.e}\onedot} 
 
\def\etc{\emph{etc}\onedot} 
\def\wrt{with respect to } 
\def\etal{\emph{et al}\onedot}

\newcounter{iictr}
\def\@iia[#1]{\setcounter{iictr}{#1}\@iib}
\def\@iib{\iii[\roman{iictr}]\xspace}
\def\ii{\@ifnextchar[{\@iia}{\stepcounter{iictr}\@iib}}
\def\iii[#1]{({\em#1\/})}
\makeatother

\def\mypar#1{\vspace{1mm}{\bf #1.}\hspace{1mm}}
\def\mytabspace{\vspace{-3mm}}
\def\myfigspace{\vspace{-3mm}}
\def\mytabcapspace{\vspace{-3mm}}
\def\myfigcapspace{\vspace{-3mm}}

\usepackage[pagebackref=true,breaklinks=true,colorlinks,bookmarks=false]{hyperref}

\newcommand{\omitme}[1]{}

\iccvfinalcopy 

\def\iccvPaperID{270} 

\ificcvfinal\fi

\begin{document}

\title{Predicting Deeper into the Future of Semantic Segmentation}

\author{
Pauline Luc$^{1,2}$\thanks{These authors contributed equally}
\hspace{7mm}
Natalia Neverova$^1$\addtocounter{footnote}{-1}\footnotemark\addtocounter{footnote}{1}
\hspace{7mm}
Camille Couprie$^1$
\hspace{7mm}
Jakob Verbeek$^2$
\hspace{7mm}
Yann LeCun$^{1,3}$\\
$^1$ Facebook AI Research\\
$^2$ Inria Grenoble, Laboratoire Jean Kuntzmann, Universit\'e Grenoble Alpes\\
$^3$ New York University\\
{\tt\small \{paulineluc,nneverova,coupriec,yann\}@fb.com \hspace{7mm} jakob.verbeek@inria.fr}
}

\maketitle
\thispagestyle{empty}


\begin{abstract}
The ability to predict and therefore to anticipate the future is an important
attribute of intelligence. It is also of utmost importance in real-time
systems, \eg in robotics or autonomous driving, which depend on visual scene
understanding for decision making. While prediction of the raw RGB pixel values
in future video frames has been studied in previous work, here we introduce the
novel task of predicting semantic segmentations of future frames.  Given a
sequence of video frames, our goal is to predict
segmentation maps of not yet observed video frames that lie up to a second or
further in the future.  We develop an autoregressive convolutional neural
network that learns to iteratively generate multiple frames.
Our results on the Cityscapes dataset show that directly predicting future segmentations is substantially
better than predicting and then segmenting future RGB frames.
Prediction results up to half a second in the future are visually convincing
and are much more accurate than those of a baseline based on warping semantic segmentations using optical flow.
\end{abstract}


\begin{figure}
\begin{center}
  \includegraphics[width=1\linewidth]{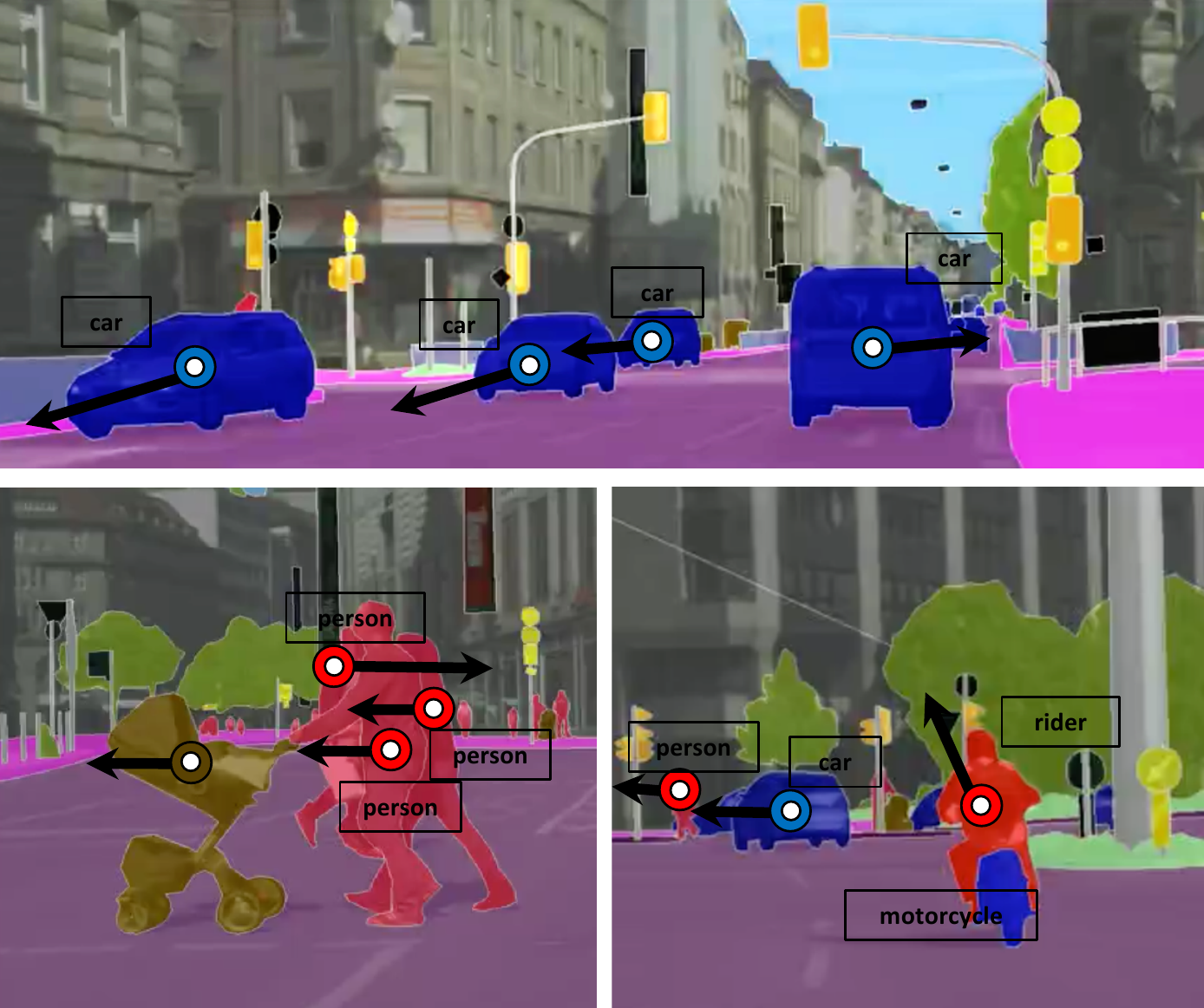}
\end{center}
\myfigcapspace
   \caption{Our models learn semantic-level scene dynamics to predict semantic segmentations of unobserved future frames given several past frames.
 }
\label{fig:teaser}
\myfigcapspace
\end{figure}

\section{Introduction}

Prediction and anticipation of future events is a key component of intelligent
decision-making \cite{sutton98book}.  Building smarter robotic systems and
autonomous vehicles implies making decisions based on the analysis of the
current situation and hypotheses made on what could happen
next~\cite{Dosovitskiy2017}.
While humans can predict vehicle or pedestrian trajectories effortlessly and at the reflex level,
it remains an open challenge for current computer vision systems.
Besides the long term goal of learning a good
representation allowing machines to reason about future events, an application
which directly benefits from our work is autonomous driving. In this domain,
approaches are either based on a number of semantic decompositions such as
road and obstacle detection,
or directly learn a mapping from visual input to driving instructions end-to-end.
Recent work from Mobileye~\cite{ShalevShwartzS16sample}
demonstrated an advantage of the semantic abstraction approach in lowering the required
amount of training data and decreasing the probability of failure. Other work
\cite{Shalev-ShwartzB16longterm} uses future prediction to facilitate long-term planning
problems and forms a direct motivation for our work.

The task of predicting future RGB video frames given preceding ones is
interesting to assess if current vision systems are able to reason about
future events, and it has recently received significant
attention~\cite{KalchbrennerOSD16videopixelnet,MathieuCouprieLeCun2016,RanzatoSzlamBruna2014,SrivastavaMS15}.
Modeling raw RGB intensities is, however, overly
complicated as compared to predicting future high-level scene properties, while
the latter is sufficient for many applications.  Such high-level future
prediction has been studied in various forms, \eg by explicitly forecasting
trajectories of people and other objects in future video
frames~\cite{Alahi_2014_CVPR,FouheyZitnick2014future,Hoai-DelaTorreIJCV14,kitani12eccv,Lan2014futureAction,Pei2011futureEvent}.
In our work we do not explicitly model objects or other scene elements, but
instead model the dynamics of semantic segmentation maps of object categories
with convolutional neural networks.
Semantic segmentation is one of the most complete forms of visual scene
understanding, where the goal is to label each pixel with the corresponding
semantic label (\eg, \emph{tree, pedestrian, car}, \etc).  In our work, we build
upon the recent progress in this area
\cite{farabet13pami,long15cvpr,chen2015iclr, YuKoltun2016, PatrauceanHandaCipolla2016, nilsson2016arxiv,lin17cvpr},
and develop models to predict the semantic segmentation of future video frames,
given several preceding frames.  See Figure~\ref{fig:teaser} for an
illustration.

The pixel-level annotations needed for semantic segmentation are expensive to
acquire, and this is even worse if we need annotations for each video frame.
To alleviate this issue
we rely on state-of-the-art semantic image segmentation models to label all
frames in videos, and then learn our future segmentation prediction models from
these automatically generated annotations.

We systematically study the effect of using RGB frames and/or segmentations
as inputs and targets for our models, and the impact of various loss
  functions.
Our experiments on the Cityscapes dataset \cite{Cordts2016Cityscapes} suggest that it is advantageous to directly predict future frames at the abstract semantic-level, rather than to predict the low-level RGB appearance of future frames and then to apply a semantic segmentation model on these.
By moving away from raw RGB predictions and modeling pixel-level object labels instead,
the network's modeling capacity seems better allocated to learn basic physics and
  object interaction dynamics.

In this work we make two contributions:
\begin{itemize}
\item We introduce the novel task of predicting future frames in the space
  of semantic segmentation.
	Compared with prediction of the RGB intensities, we show that we can predict further into the future, and hence model more interesting distributions.

\item We propose an autoregressive model which convincingly predicts segmentations up to 0.5 seconds into the future. The
    mean IoU of our predictions reaches two thirds of the ones obtained by
	the method used to automatically generate the dense video annotations used for training \cite{YuKoltun2016}.
\end{itemize}

Our approach does not require extremely costly temporally dense video annotation and its genericity allows different architectures for still-image segmentation and future segmentation prediction to be swapped in.

\section{Related work}
\label{ref:relatedwork}

Here we discuss the most relevant related work on video forecasting and on
disambiguating learning under uncertainty, in particular using adversarial training.

\mypar{Video forecasting} Several authors developed methods related to our work to improve the
temporal stability of semantic video segmentation.  Jin \etal
\cite{Jin2016VideoParsingPredictive} train a model to predict the semantic
segmentation of the immediate next image from the preceding input frames, and
fuse this prediction with the segmentation computed from the next input frame.
Nilsson and Sminchisescu~\cite{nilsson2016arxiv} use a convolutional RNN model
with a spatial transformer component \cite{jaderberg15nips} to accumulate the
information from past and future frames in order to improve prediction of the
current frame segmentation.  In a similar spirit, Patraucean \etal
~\cite{PatrauceanHandaCipolla2016} employ a convolutional RNN to implicitly
predict the optical flow, and use these to warp and aggregate per-frame
segmentations.  In contrast, our work is focused on predicting future
segmentations without seeing the corresponding frames. Most importantly, we
target a longer time horizon than a single frame.

A second line of related work focuses on generative models for future video
frame forecasting. Ranzato \etal~\cite{RanzatoSzlamBruna2014} introduced the
first baseline of next video frame prediction.  Srivastava \etal
\cite{SrivastavaMS15} developed a Long Short Term Memory (LSTM)
\cite{hochreiter1997neco} architecture for the task, and demonstrated a gain in
action classification using the learned features.  Mathieu \etal
\cite{MathieuCouprieLeCun2016} improved the predictions using a multi-scale
convolutional architecture, adversarial training
\cite{GoodfellowPougetMirza2014}, and a gradient difference loss. A similar
training strategy was employed for future frame predictions in time-lapse videos \cite{Zhou2016TimeLapse}. To reduce the
number of parameters to estimate, several authors reparameterize the problem to
predict frame transformations instead of raw
pixels~\cite{FinnGL16,AmersfoortKannanRanzato2016}.  Luo \etal
\cite{Luo2017unsupervised} employ a convolutional LSTM architecture to predict
sequences of up to eight frames of optical flow in RGB-D videos.  The video pixel
network of Kalchbrenner \etal~\cite{KalchbrennerOSD16videopixelnet} combine
LSTMs for temporal modeling with spatial autoregressive modeling.  Rather than
predicting pixels or flows, Vondrick \etal~\cite{VondrickPirsiavashTorralba2015}
instead predict features in future frames. They predict the activations of the last hidden layer of AlexNet \cite{krizhevsky12nips} in future frames, and use these to anticipate objects and actions.

\mypar{Learning under uncertainty} Generative adversarial networks
(GANs)~\cite{GoodfellowPougetMirza2014} and variational autoencoders
(VAEs)~\cite{kingma14iclr} are deep latent variable models that can be
used to deal with the inherent uncertainty in future-prediction tasks.
An interesting approach using GANs for unsupervised image representation learning was simultaneously proposed in \cite{DonahueKrahenbuhlDarrell2016} and \cite{dumoulin17iclr}, where the generative model is trained along with an inference model that maps images to their latent representations.
Vondrick \etal \cite{Vondrick2016GeneratingVideos} showed that GANs can be
applied to video generation. They use a two-stream generative model: one stream
generates a static background, while the other generates a dynamic foreground
sequence which is pasted on the background. Yang \etal \cite{yang17iclr} use
similar ideas to develop an iterative image generation model where objects are
sequentially pasted on the image canvas using a recurrent GAN.  Xue
\etal~\cite{xue16nips} predict future video frames from a single given frame
using a VAE approach.  Similarly, Walker \etal~\cite{WalkerDoerschGupta2016}
perform forecasting with a VAE, predicting feature point trajectories from still
images.

\section{Predicting future frames and segmentations}
\label{sec:architecture}

We start by presenting different scenarios to predict RGB pixel values and/or segmentations of the next video frame.
In Section \ref{sec:autor} we describe two extensions of the single-frame prediction model to predict further into the future.

\omitme{
\begin{table}
\begin{center}
\begin{tabular}{lcc}
\toprule
Model & Input & Output\\
\midrule
\model{X}{X} 	&  $X_{1:t}$ 			& $X_{t+1}$\\
\model{S}{S} 	&  $S_{1:t}$ 			& $S_{t+1}$\\
\model{XS}{X}	&  $(X_{1:t}, S_{1:t})$ & $X_{t+1}$ 	\\
\model{XS}{S} 	&  $(X_{1:t}, S_{1:t})$ & $S_{t+1}$ 	\\
\model{XS}{XS} 	&  $(X_{1:t}, S_{1:t})$ & $(X_{t+1},S_{t+1})$ \\
\bottomrule
\end{tabular}
\end{center}
\mytabcapspace
\caption{The 5 models for single-frame prediction.}
\label{tab:models}
\mytabspace
\end{table}}

\subsection{Single-frame prediction models}
\label{sec:models}

Pixel-level supervision is laborious to acquire for semantic image segmentation, and even more so for its video counterpart.
To circumvent the need
for datasets with per-frame annotations, we use the state-of-the-art Dilation10
semantic image segmentation network \cite{YuKoltun2016} to provide input and target
semantic segmentations for all frames in each video. We use the resulting
temporally dense segmentation sequences to learn our models.

Let us denote with $X_i$ the $i$-th frame of a video sequence and denote the
sequence of frames from $X_t$ to $X_T$ as $X_{t:T}$.  We denote by $S_i$
the semantic segmentation of frame $X_i$ given the Dilation10 network.
We represent the segmentations $S_i$ using the final softmax layer's pre-activations, rather than the probabilities it produces.
This is motivated by recent
observations in network distillation that the softmax pre-activations carry more
information 
\cite{ba14nips,hinton14dlws}.
For single-frame future prediction, we consider five different models that
differ in whether they take RGB frames and/or segmentations as their inputs and
targets: model \model{X}{X} takes  $X_{1:t}$ and predicts $X_{t+1}$, model
  \model{S}{S} takes $S_{1:t}$ and predicts $S_{t+1}$, models \model{XS}{X} and
  \model{XS}{S} take $(X_{1:t}, S_{1:t})$ and predict respectively   $X_{t+1}$
  and  $S_{t+1}$, and finally model \model{XS}{XS} takes $(X_{1:t}, S_{1:t})$
  and predicts $(X_{t+1}, S_{t+1})$.


\begin{figure}
  \begin{center}
    \includegraphics[width=0.3\textwidth]{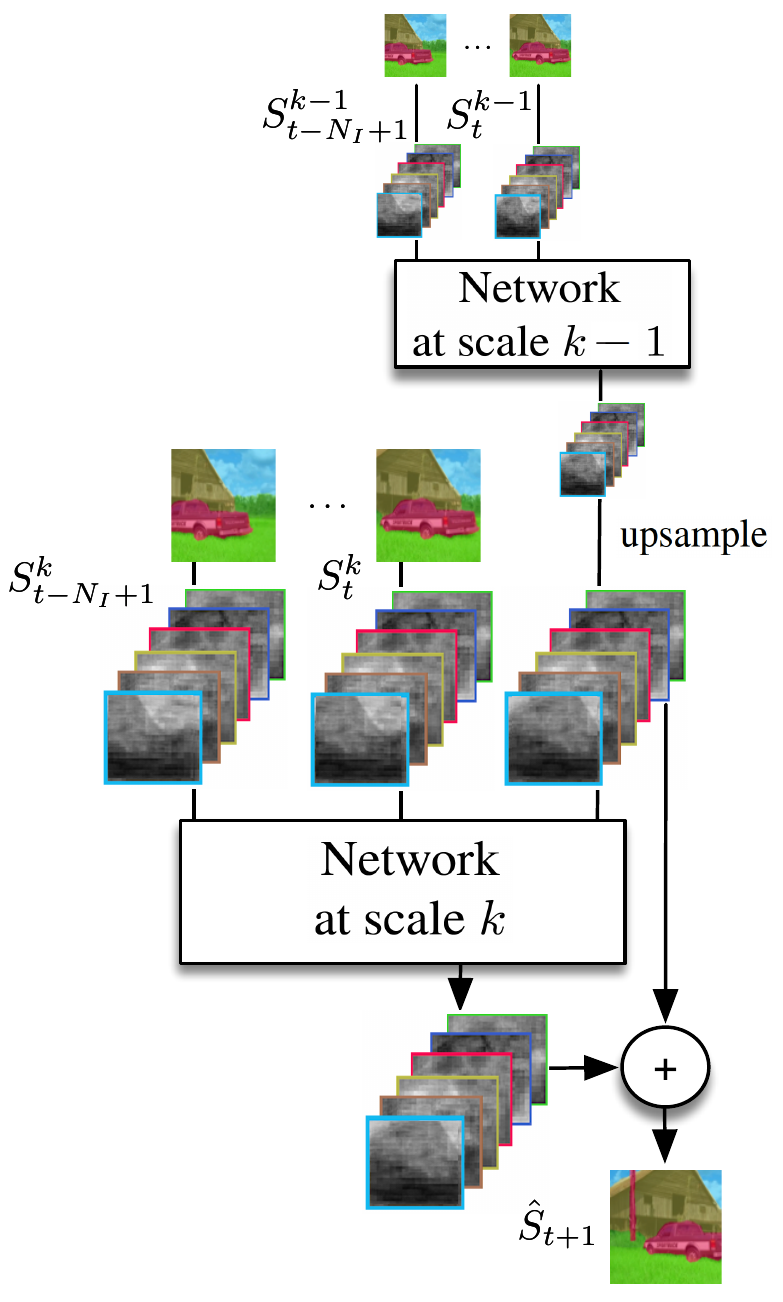} \\
    \end{center}
    \myfigcapspace
      \caption{Multi-scale architecture of the \model{S}{S} model that predicts
        the semantic segmentation of the next frame given the segmentation maps
        of the $N_I$ previous frames.}
\label{fig:multiscale}
    \myfigspace
\end{figure}

\mypar{Architectures}
Model \model{X}{X} is a next frame prediction model, for which we use the multi-scale
network of Mathieu \etal \cite{MathieuCouprieLeCun2016} with two spatial
scales. Noting $C$ the number of output channels,
each scale module is  a four-layer convolutional network alternating
convolutions and ReLU operations, outputting feature maps with 128, 256, 128, $C$  channels each,
and filters of size 3 for the smaller scale, and 5, 3, 3, 5 for the
larger scale. The last non-linear function is a
hyperbolic tangent, to ensure that the predicted RGB values lie in the range $[-1, 1]$.
The output at a coarser scale is upsampled, and used in input to the next scale module
together with a copy of the input at that scale.

For models that predict segmentations $S_{t+1}$, we removed the last hyperbolic
tangent non-linearities for the corresponding output channels, since the softmax pre-activations are not limited to a fixed range.  Apart from this difference,
the \model{S}{S} model, that predicts the next segmentation from past ones,
has the same architecture as the \model{X}{X} model.

The multi-scale architecture of the \model{S}{S} model is illustrated in
Figure~\ref{fig:multiscale}.  The other models (\model{XS}{X}, \model{XS}{S},
and \model{XS}{XS}), which take both RGB frames and segmentation maps as input,
also use the same internal architecture.

\mypar{Loss function}
Following \cite{MathieuCouprieLeCun2016}, for all models, the loss function between the model output $\hat{Y}$ and
the target output $Y$ is the sum of an $\ell_1$ loss and a gradient difference
loss: \bea \mathcal{L}(\hat{Y},Y) = \mathcal{L}_{\ell_1}(\hat{Y},Y) +
\mathcal{L}_{\textrm{gdl}}(\hat{Y},Y).  \eea Using $Y_{ij}$ to denote the pixel
elements in $Y$, and similarly for $\hat{Y}$, the losses are defined as: \bea
\mathcal{L}_{\ell_1}(\hat{Y},Y) & = & \sum_{i,j} \big|
Y_{ij}-\hat{Y}_{ij}\big|,\\ \mathcal{L}_{\textrm{gdl}}(\hat{Y},Y) & = &
\sum_{i,j} \Big| \big|Y_{i,j}-Y_{i-1,j}\big| -
\big|\hat{Y}_{i,j}-\hat{Y}_{i-1,j}\big| \Big| \nonumber \\ && + \Big|
\big|Y_{i,j-1}-Y_{i,j}\big| - \big|\hat{Y}_{i,j-1}-\hat{Y}_{i,j}\big| \Big|,
\eea where $|\cdot|$ denotes the absolute value function.
The $\ell_1$ loss tries to match all pixel predictions independently to their corresponding target values.
The gradient difference loss, instead, penalizes errors in the gradients of the prediction.
This loss is relatively insensitive to low-frequency mismatches between prediction and target
(\eg, adding a constant to all pixels does not affect the loss),
and is more sensitive to high-frequency mismatches that are perceptually more significant (\eg errors along the contours of an object).
We present a comparison of this loss with a multiclass cross entropy loss in Section~\ref{sec:experiments}.

\mypar{Adversarial training}
As shown by Mathieu \etal~\cite{MathieuCouprieLeCun2016} in the context of raw
images, introducing an adversarial loss allows the model to disambiguate
between modes corresponding to different turns of events, and reduces blur
associated with this uncertainty. Luc \etal~\cite{luc16nips}
 demonstrated the positive influence of adversarial training for semantic
image segmentation.

Our formulation of the adversarial loss term is based on the 
recently introduced Wasserstein GAN~\cite{ArjovskyCB17}, with some modifications
for the semantic segmentation application.  In the case of the \model{S}{S}
model, the parameters $\theta$ of the discriminator $\mathcal{D}_\theta$ are
trained to maximize the absolute difference between its output 
for ground truth sequences
$(S_{1:t}, S_{t+1})$ and sequences $(S_{1:t}, \hat{S}_{t+1})$ predicted by our
model:
\bea
\max_{\theta} \left| \sigma\left(\mathcal{D}_\theta\left(S_{1:t},S_{t+1}\right)\right) -
\sigma\left(\mathcal{D}_\theta\left(S_{1:t},\hat{S}_{t+1}\right)\right)\right|.
\eea

The outputs
produced by the predictive model are softmax pre-activation maps with unbounded
values. In the Wasserstein GAN they are encouraged to grow indefinitely. To
avoid this and stabilize training, we employ an additional sigmoid non-linearity
$\sigma$ at the output of the discriminator, and set explicit targets for two
kinds of outputs: $0$ for generated sequences and $\alpha$ for real training
sequences, set to $0.9$ to prevent saturation.

The adversarial regularization term for our predictive model (\ie the ``generator'') then takes the following form:
\bea
\mathcal{L}_{\text{adv}}(S_{1:t},\hat{S}_{t+1}) & = & \lambda \left|\sigma\left(\mathcal{D}_\theta\left(S_{1:t},\hat{S}_{t+1}\right)\right)-\alpha\right|.
\eea

The structure of the discriminator network is derived from the two-scale
architecture described above. Additional details are provided in the
supplementary material.

\begin{figure}
\begin{center}
\includegraphics[scale=.52]{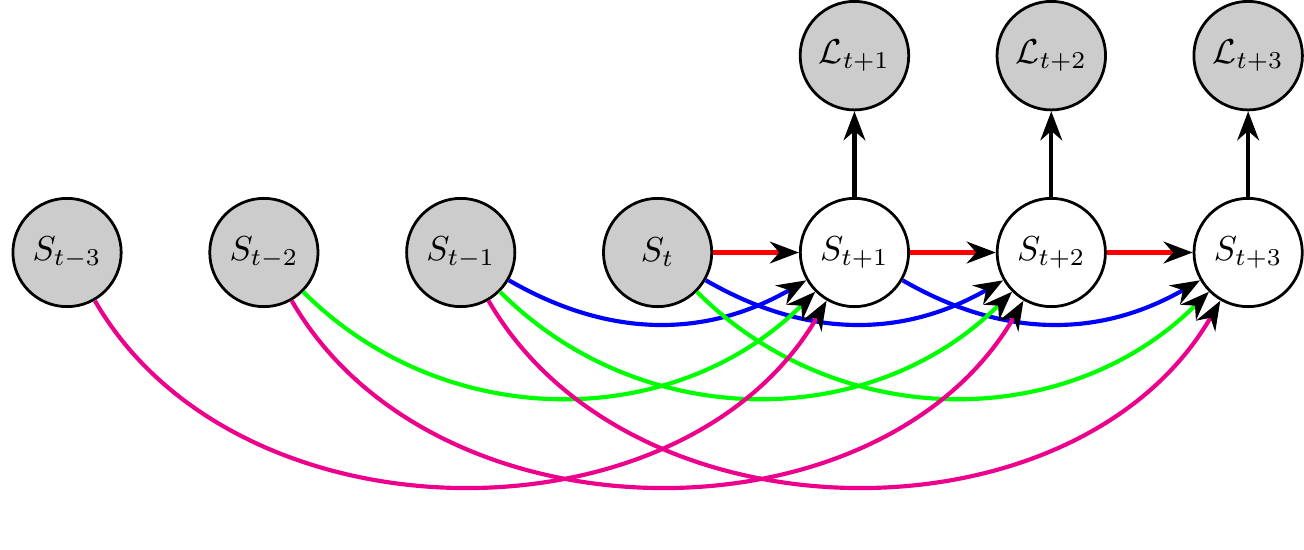}
\includegraphics[scale=.52]{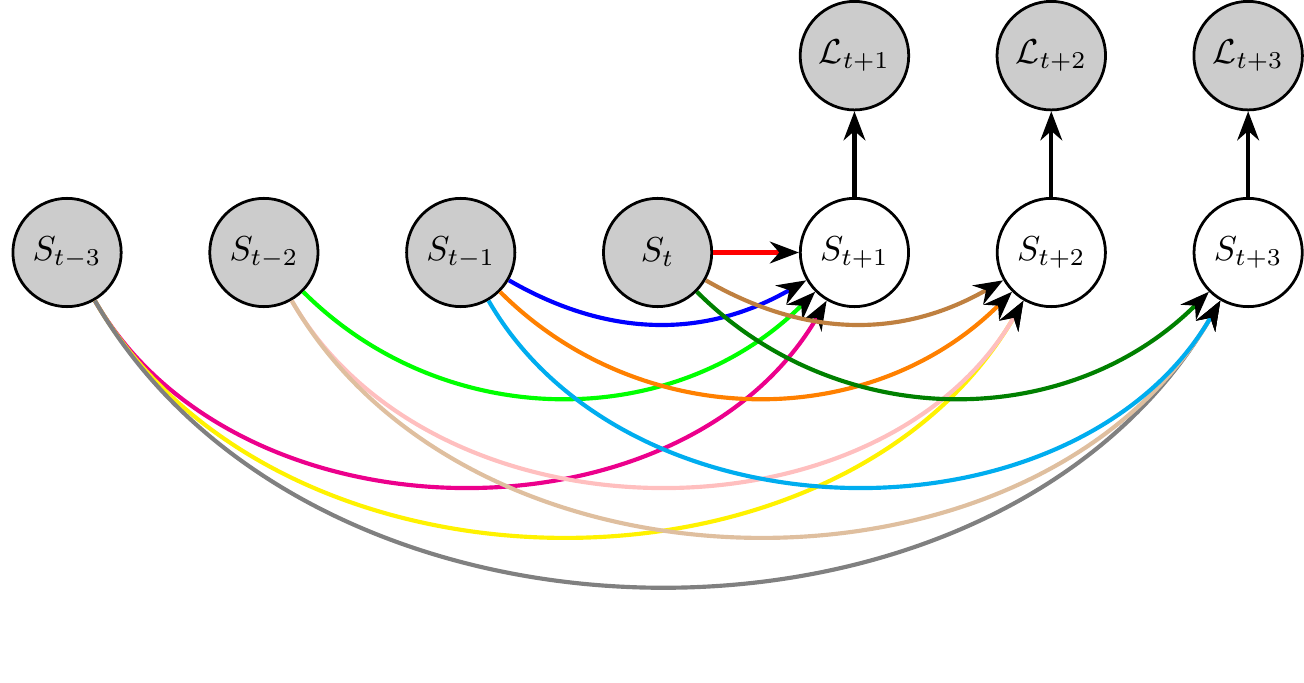}
\end{center}
\myfigcapspace
\caption{Illustration of the \omitme{for single time-step prediction
    (top), and extensions for multiple time-step prediction: the}
  autoregressive (top) and batch (bottom) models. The autoregressive model shares parameters over time; dependency links are colored accordingly.
}
\label{fig:recursive}
\myfigspace
\end{figure}

\subsection{Predicting deeper into the future}
\label{sec:autor}

We consider two extensions of  the previous models to predict further into the future than a single frame.
The first is to expand the output of the network to comprise a batch of $m$ frames,
\ie to output $X_{t+1:t+m}$ and/or $S_{t+1:t+m}$.
We refer to this as the ``batch'' approach.
The  drawback of this approach is that it ignores the recurrent structure of the problem.
That is, it ignores the fact that $S_{t+1}$ depends on $S_{1:t}$ in the same manner as $S_{t+2}$ depends on $S_{2:t+1}$.
As a result, the capacity of the model is split to predict the $m$ output frames, and the number of parameters in the last layer scales linearly with the number of output frames.

In our second approach, we leverage the recurrence property, and iteratively
apply a model that predicts a single step into the future, using its
prediction for time $t+1$ as an input to predict at time $t+2$, and so on.  This allows us
to predict arbitrarily far into the future in an autoregressive manner, without
resources scaling with the number of time-steps we want to predict. We
refer to this approach as ``autoregressive''.  See Figure~\ref{fig:recursive} for a
schematic illustration of the two extensions for
multiple time-step predictions.

In the autoregressive mode, we first evaluate the models trained for single-frame prediction, then we fine-tune these models using backpropagation through time \cite{werbos1988nn}, to account for the fact that mistakes at each time-step affect all later time-steps. 

\omitme{
In the autoregressive mode, we either use the models as trained to predict one
time-step ahead, or we fine-tune these models by taking into account
the impact the autoregressive approach has on predictions farther away than a single frame.
In the latter case, during training we first make a forward pass, predicting one frame ahead
at a time, and using the most recent outputs to predict the next time step.  We
then back-propagate the gradients through time \cite{werbos1988nn}, where the
gradients \wrt the prediction at time $t$ are based on the loss at time $t$, and
the impact on the losses at later time steps.}

\section{Experiments}
\label{sec:experiments}

Before presenting our experimental results, we first describe the dataset and evaluation metrics in \sect{dataset}.
We then present results on short-term (i.e. single-frame) prediction, mid-term prediction (0.5 sec.), and long-term prediction (10 sec.).

\renewcommand\arraystretch{1.1}
\renewcommand\tabcolsep{1pt}
\begin{figure}
  \begin{center}
    \def\myfig#1{\includegraphics[width=0.33\linewidth]{short-term-7-#1}}
    \begin{tabular}{ccc}
       \myfig{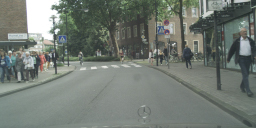} &        \myfig{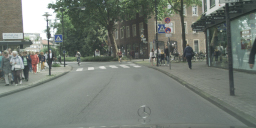} &        \myfig{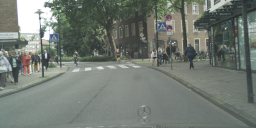} \\
           Input: $X_{t-3}$  &  Input: $X_{t}$ &        Ground truth: $X_{t+3}$\\
      \myfig{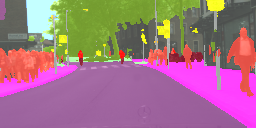}& \myfig{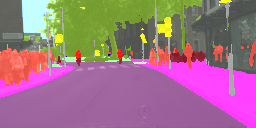}  &\myfig{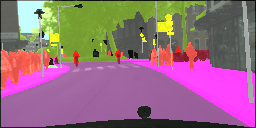}\\
        Input: $S_{t-3}$  &  Input: $S_t$ &         Ground truth:  $S_{t+3}$ \\
      \myfig{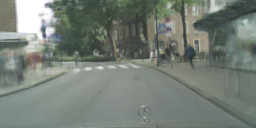}& \myfig{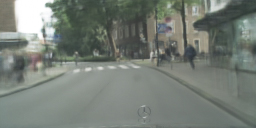}& \myfig{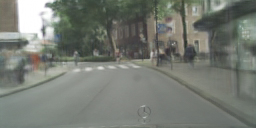}\\
\model{X}{X}: $X_{t+3}$  &  \model{XS}{X}: $X_{t+3}$ &   \model{XS}{XS}:       $X_{t+3}$\\
       \myfig{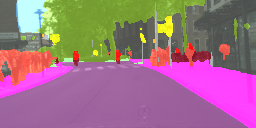}			& \myfig{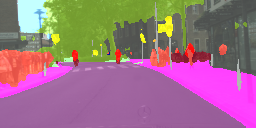}			& \myfig{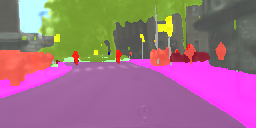}\\
\model{S}{S}: $S_{t+3}$	& \model{XS}{S}: $S_{t+3}$	& \model{XS}{XS}:       $S_{t+3}$\\
       \myfig{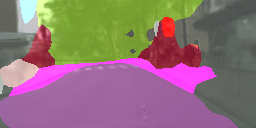} 	& \myfig{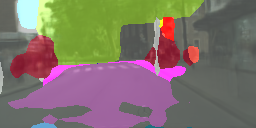} 	& \myfig{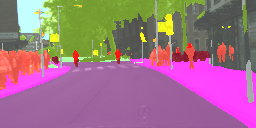}\\
\model{X}{X}: $S_{t+3}$  		& \model{XS}{X}: $S_{t+3}$ 		& Dilation10: $S_{t+3}$
         \end{tabular}
    \end{center}
    \myfigcapspace
   \caption{Short-term predictions of RGB frame $X_{t+3}$ and segmentation $S_{t+3}$  using our different models, compared to ground truth, and Dilation10 oracle that has seen $X_{t+3}$.
    }
\label{fig:shortterm}
    \myfigspace
\end{figure}
\renewcommand\tabcolsep{3pt}

\subsection{Dataset and evaluation metrics}
\label{sec:dataset}

The Cityscapes dataset \cite{Cordts2016Cityscapes} contains 2,975 training, 500
validation and 1,525 testing video sequences of 1.8 second. Each sequence
consists of 30 frames, and a ground-truth semantic segmentation is available for
the 20-th frame.  The segmentation outputs of the Dilation10 network
\cite{YuKoltun2016} are produced at a resolution of $128\times 256$ and we
perform all experiments at this resolution. For this purpose, we also downsample
RGB frames and ground truth to this resolution.  We report performance of our
models on the Cityscapes validation set, and refer to the supplementary material
for results on the test set.

We assess performance using the standard mean Intersection over Union (IoU)
measure, computed \wrt the ground truth segmentation of the 20-th frame in each
sequence (IoU GT).  We also compute the IoU measure \wrt the segmentation
produced using the Dilation10 network~\cite{YuKoltun2016} for the 20-th frame
(IoU SEG).  The IoU SEG metric allows us to validate
our models \wrt the target segmentations from which they are trained.  Finally,
we compute the mean IoU across categories that can move in the scene:
\emph{person, rider, car, truck, bus, train, motorcycle, and bicycle} (IoU-MO,
for ``moving objects'').

To evaluate the quality of the frame RGB predictions, we compute the Peak Signal to
Noise Ratio (PSNR) and the Structural Similarity Index Measure (SSIM) measures
\cite{Wang04SSIM}. The SSIM measures similarity
between two images, ranging between $-1$ for very dissimilar inputs to $+1$ when
the inputs are the same.
It is based on comparing local patterns of pixel intensities normalized
for luminance and contrast.

Unless specified otherwise, we train our models using a frame interval of 3, and taking 4 frames and/or segmentations as input.
That is, the input sequence consists of frames $\{X_{t-9}, X_{t-6}, X_{t-3}, X_{t}\}$, and similarly for segmentations.
We performed patch-wise training with $64\times64$ patches for the largest scale resolution,
enabling equal class frequency sampling as in \cite{farabet13pami}, using mini-batches of four patches and a learning rate of 0.01.

\subsection{Short-term prediction}
\label{sec:short}

In our first experiment, we compare the five different input-output
representations.  For models that do not directly predict future segmentations,
we generate segmentations using the Dilation10 network based on the predicted
RGB frames.  We also include two baselines.  The first baseline copies the last
input frame to the output.  The second baseline estimates the optical flow
between the last two inputs, and warps the last input using the estimated
flow. Further details are given in the supplementary material.  Comparison with
tracking-based approaches is difficult since (i) segmentation is performed
densely and lacks the notion of object instances used by object trackers, and
(ii) ``stuff'' categories (road, vegetation, \etc), useful for drivable area
detection in the context of autonomous driving, are not suitable for modeling
with tracking-based approaches.

\begin{table}
{\small
\begin{center}
  \begin{tabular}{lcccc}
    \toprule
    Method  & PSNR & SSIM & IoU  & IoU-MO \\
     &  &  & GT (SEG) &  GT (SEG) \\
\midrule
   Copy  last input &  20.6 & 0.65  & 49.4 (54.6) & 43.4 (48.2) \\
   Warp last input & 23.7 & 0.76 & 59.0 ({\bf 67.3}) & 54.4  ({\bf 63.3}) \\ 
    \midrule
    Model \model{X}{X}   & 24.0 & \bf0.77     & 23.0 (22.3) & 12.8 (11.4) \\
    Model  \model{S}{S}    &   ---  &    --- &  58.3 (64.9) & 53.8 (59.8) \\

   Model   \model{XS}{X}    & \bf24.2 & \bf0.77 & 22.4 (22.5) & 10.8 (10.0) \\
   Model   \model{XS}{S}    & ---  &   ---         & 58.2 (64.6) & 53.7 (59.9) \\
   Model  \model{XS}{XS}    & 24.0   & 0.76   & 55.5 (61.1) & 50.7 (55.8) \\
   \midrule
    Model  \model{S}{S}-adv. &       ---  &     --- &  58.3 (65.0) & 53.9 (60.2) \\
   Model \model{S}{S}-dil & ---  &     --- & {\bf 59.4} (66.8) & {\bf 55.3} (63.0) \\
    \bottomrule
  \end{tabular}
  \end{center}
  }
  \mytabcapspace
  \caption{Short-term prediction accuracy of baselines and of our models taking either RGB
    frames (X) and/or segmentations (S) as input and output.  For reference: the 59.4 IoU corresponds to 91.8\% per pixel  accuracy.}
  \label{tab:shortcityscapes}
    \mytabspace
\end{table}

In Figure~\ref{fig:shortterm}, we show qualitative results of the predictions for one of the validation sequences.
From the quantitative result in
\tab{shortcityscapes} we make several observations.
First, in terms of RGB frame prediction (PSNR and SSIM), the performance is comparable for the three models \model{X}{X}, \model{XS}{X}, and \model{XS}{XS}, and improves over the two baselines.
This shows that our models learn non-trivial scene dynamics in the RGB pixel space,
and that adding semantic segmentations either at input and/or output does not have a substantial impact on this ability.

Second, in terms of the IoU segmentation metrics, the models that directly predict future segmentations (\model{S}{S}, \model{XS}{S}, \model{XS}{XS}) perform much better than the models that only predict the RGB frames.
This suggests that artifacts in the RGB frame predictions degrade the performance of the Dilation10 network. See also the corresponding RGB frame predictions in Figure~\ref{fig:shortterm}.

Third, the \model{XS}{XS} model, which predicts both segmentations and RGB frames performs somewhat worse than the models that only predict segmentations (\model{S}{S} and \model{XS}{S}), suggesting that some of the modeling capacity is compromised by jointly predicting the RGB frames.

Fourth, we find that fine-tuning the \model{S}{S} model using adversarial training
(\model{S}{S}-adv) does not lead to a significant improvement over  normal training.

Table~\ref{tbl:abl} presents results of an ablation study of the \model{S}{S} model,
assessing the impact of the different loss functions, as well as the impact of using one or two scales.
We include the results obtained using the Dilation10 model as an ``oracle'', that predicts the future segmentation based on the future RGB frame, which is not accessible to our other models.
This oracle result gives the maximum performance that could be expected, since this oracle was used to provide the training data - we can thus only expect our models to have at best comparable performance with this oracle.
  All variants of the \model{S}{S} model were trained during about 960,000 iterations, taking about four days of training on a single GPU.
The results show that using two scales improves the performance, as does the addition of the gradient difference loss.
Training with the  $\ell_1$ and/or gdl loss on the softmax pre-activations gives better results as compared to training using the
multi-class cross-entropy (MCE) loss on the segmentation labels.
This is in line with observations made in network distillation~\cite{ba14nips,hinton14dlws}.

Finally, we perform further architecture exploration for the \model{S}{S}
  model, which performed best. We propose a simpler, deeper, and more efficient
  architecture with dilated convolutions~\cite{YuKoltun2016}, to expand the field of view while retaining accurate localization for the predictions. We call this
  model \model{S}{S}-dil, and provide details in the supplementary material.
This model gives best overall results, reported in Table \ref{tab:shortcityscapes}.

\begin{table}
{\small
  \begin{center}
    \begin{tabular}{lccc}
      \toprule
      Model & IoU GT & IoU SEG & IoU-MO  GT \\
    \midrule
       Dilation10 oracle & 68.8 & 100  & 64.7 \\
       \midrule
       \model{S}{S}, 2 scales, $\ell_1$+gdl  &  \bf 58.3&  \bf 64.9 & \bf 53.8 \\
\model{S}{S}, 1 scale, $\ell_1$+gdl  & 57.7 & 63.9 & 52.6 \\
\model{S}{S}, 2 scales, $\ell_1$& 57.6 & 64.0 & 53.2 \\
\model{S}{S}, 2 scales, MCE &  55.5 & 60.9& 49.7\\
\bottomrule
  \end{tabular}
  \end{center}
  }
  \mytabcapspace
  \caption{Ablation study with the \model{S}{S} model, and comparison to a Dilation10 oracle that predicts the future segmentation using the future RGB frame as input.
    }
  \label{tbl:abl}
    \mytabspace
  \end{table}

\begin{figure*}
  \begin{center}
    \begin{tabular}{cc}
      \renewcommand\arraystretch{1}
\renewcommand\tabcolsep{1pt}
       \begin{subfigure}[t]{0.49\textwidth}
         \begin{tabular}{cc}

           \includegraphics[width=0.48\linewidth]{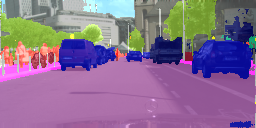}&
           \includegraphics[width=0.48\linewidth]{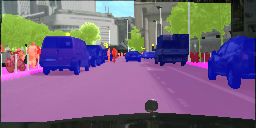}\\
              $X_{t}, S_{t}$ & $X_{t+9}$, GT \\
           \includegraphics[width=0.48\linewidth]{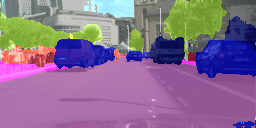}&
           \includegraphics[width=0.48\linewidth]{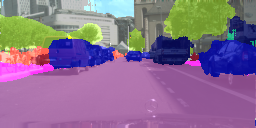}\\
            Batch predictions at $t+3$ & at $t+9$ \\
            \includegraphics[width=0.48\linewidth]{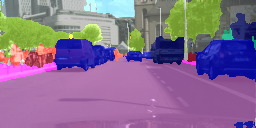}&
            \includegraphics[width=0.48\linewidth]{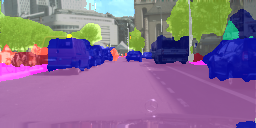}\\
            Autoregressive pred. at $t+3$ & at $t+9$ \\
             \includegraphics[width=0.48\linewidth]{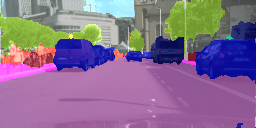}&
            \includegraphics[width=0.48\linewidth]{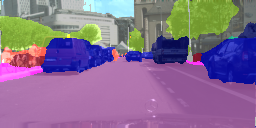}\\
            AR fine-tune  pred.\ at $t+3$ & at $t+9$ \\
         \end{tabular}\\
       \end{subfigure}&
        \renewcommand\arraystretch{1}
\renewcommand\tabcolsep{1pt}
       \begin{subfigure}[t]{0.49\textwidth}
      \begin{tabular}{cc}
          \includegraphics[width=0.48\linewidth]{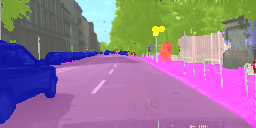}&
           \includegraphics[width=0.48\linewidth]{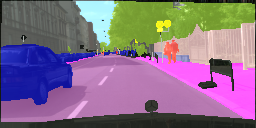}\\
              $X_{t}, S_{t}$ & $X_{t+9}$, GT \\
           \includegraphics[width=0.48\linewidth]{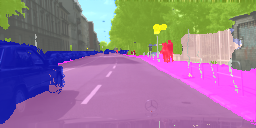}&
           \includegraphics[width=0.48\linewidth]{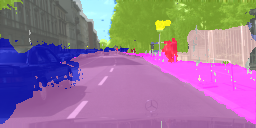}\\
           Optical flow at $t+3$ & at $t+9$ \\
            \includegraphics[width=0.48\linewidth]{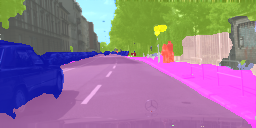}&
           \includegraphics[width=0.48\linewidth]{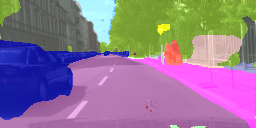}\\
            AR adv.\ pred.\  at $t+3$ & at $t+9$ \\
            \includegraphics[width=0.48\linewidth]{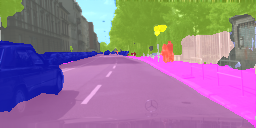}&
           \includegraphics[width=0.48\linewidth]{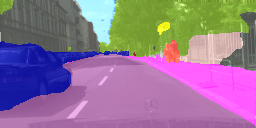}\\
            AR fine-tune pred. at $t+3$ & at $t+9$ \\
         \end{tabular}\\
       \end{subfigure}
       \end{tabular}
    \end{center}
        \myfigcapspace
   \caption{Optical flow baseline, \model{S}{S} autoregressive and \model{S}{S} batch predictions for two sequences (first sequence left, second sequence right).
     First row: last input and  ground truth. Other rows show predictions
     overlayed with the true future frames. The full results are provided in the
     supplementary material.
      }
\label{fig:midterm_model1}
    \myfigspace
\end{figure*}

\renewcommand\tabcolsep{3pt}

\subsection{Mid-term prediction}
\label{sec:mid}

We now address the more challenging task of predicting the mid-term future, \ie
the next 0.5 second.  In these experiments we take in input frames 2, 5, 8,
and 11, and predict outputs for frames 14, 17 and 20.  We compare different
strategies: batch models, autoregressive models (AR), and models
with autoregressive fine-tuning (AR fine-tune). We
  compare these strategies to our two baselines consisting in copying the last
  input, and the second one relying on optical flow. For the optical flow
  baseline, after the first prediction, we also warp the flow field so that the flow is
  applied to the correct locations at the next time-step, and so on.
   Qualitative prediction results are shown in Figure~\ref{fig:midterm_model1}. For models
\model{XS}{X} and \model{XS}{S}, the autoregressive mode is not
used because either the frame or the segmentation input are missing for
predicting from the second output on.

\begin{table}
{\small
  \begin{center}
\begin{tabular}{lcccc}
\toprule
& \multicolumn{2}{c}{Frame 14} & \multicolumn{2}{c}{Frame 20} \\
Model & PSNR & SSIM & PSNR & SSIM  \\
\midrule
Copy  last input & 20.4 & 0.64 & 18.0 & 0.55 \\
Warp last input  & 23.5 & \bf 0.76 & 19.4 & 0.59 \\ 
\hline
\model{X}{X}, AR & {\bf 23.9} & \bf 0.76 & 19.2 & 0.61\\
\model{XS}{XS}, AR  & 23.8 & \bf 0.76 & 19.3 & 0.61  \\
\model{X}{X}, batch & 23.8 & \bf 0.76 & 20.6 & \bf 0.65  \\
\model{XS}{X}, batch & {\bf 23.9} &  \bf 0.76 & {\bf 20.7} &  \bf 0.65 \\
\model{XS}{XS}, batch & 23.8 & \bf 0.76  & \bf 20.7 & 0.64 \\
\bottomrule
\end{tabular}
\end{center}
}
\mytabcapspace
  \caption{Mid-term RGB frame prediction results for frame 20 using different models in batch and autoregressive mode.}
   \label{tab:mid_img}
   \mytabspace
\end{table}

\begin{table}
  ~\\
{\small
\begin{center}
\begin{tabular}{lccc}
\toprule
Model & IoU GT & IoU SEG & IoU-MO GT\\ 
\midrule
Copy last input & 36.9  & 39.2 & 26.8 \\%
Warp last input & 44.3 & 47.2 & 37.0  \\
\midrule
\model{S}{S}, AR &  45.3  & 47.2 & 36.4\\ 
\model{S}{S}, AR, fine-tune & 46.7 & 49.7 & 39.3  \\
\model{XS}{XS}, AR  & 39.3 & 40.8 & 27.4 \\ 
\model{S}{S}, batch & 42.1 & 44.2 & 32.8\\ 
\model{XS}{S}, batch & 42.3 & 44.6 & 33.1 \\
\model{XS}{XS}, batch & 41.2 & 43.5 & 31.4 \\
\midrule
\model{S}{S}-adv, AR & 45.1 & 47.2 & 37.3  \\
\model{S}{S}-dil, AR & 46.5 & 48.6 & 38.8  \\%
\model{S}{S}-dil, AR, fine-tune & \bf47.8 &  \bf50.4 &   \bf40.8 \\
\bottomrule
\end{tabular}
\end{center}
}
\mytabcapspace
\caption{Mid-term segmentation prediction results. For reference: the 47.8  IoU corresponds to 87.9\%  per pixel accuracy.}
   \label{tab:midseg}
   \mytabspace
\end{table}

The results for RGB frame prediction in \tab{mid_img} show that for frame 14,
all models give comparable results, consistently improve over the copy baseline and perform somewhat better than the warping baseline.
For frame 20, the batch models perform best.  On the contrary, when predicting segmentations, we find that the autoregressive models perform better than the batch
models, as reported in \tab{midseg}. This is probably due to the fact that the single-step predictions are
more accurate for segmentation, which makes them more suitable for
autoregressive modeling. For RGB frame prediction, errors accumulate quickly,
leading to degraded autoregressive predictions.  Among the batch models, using
the images as input (\model{XS}{S} model) slightly helps. Predicting both the
images and segmentation (\model{XS}{XS} model) performs worst, the image
prediction task presumably taking up resources otherwise available for modeling the dynamics of the sequence.

\begin{figure*}
\def\myfig#1{\includegraphics[width=0.19\linewidth]{#1}}
\begin{center}
      \begin{tabular}{ccccccccccc}
\myfig{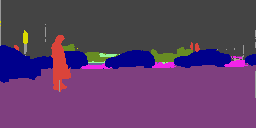}
\myfig{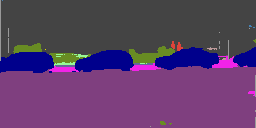}
\myfig{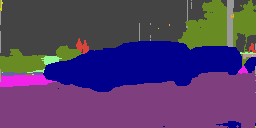}
\myfig{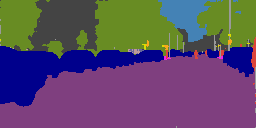}
\myfig{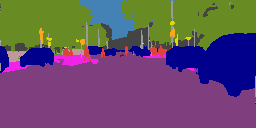}
  \end{tabular}\\
    \begin{tabular}{ccccccccccc}
\myfig{pred_4}
\myfig{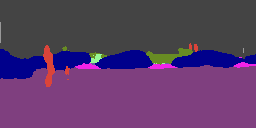}
\myfig{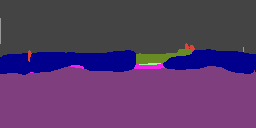}
\myfig{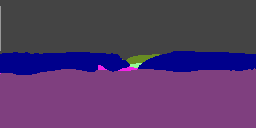}
\myfig{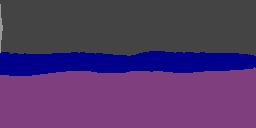}
 \end{tabular}
  \end{center}
  \myfigcapspace
   \caption{
  Last input segmentation, and ground truth segmentations at 1, 4, 7, and 10 seconds into the future (top row), and corresponding predictions of the autoregressive \model{S}{S} model trained with fine-tuning (bottom row).
     }
\label{fig:long}
  \myfigspace
\end{figure*}

Model \model{S}{S} is the most effective, as it can be applied
  in autoregressive mode, and outperforms \model{XS}{XS} in this  setting.
  In Figure~\ref{fig:midterm_model1} we compare different
  versions of this model. Visually, the first sequence shows some improvements
  using the autoregressive fine-tuned model, by more accurately matching contours of the moving cars than
  the other strategies. The second sequence displays typical
  failures of the optical flow baseline, where certain values cannot be estimated because they correspond to points that were not present in the input, \eg those at the back of the incoming car, and must be filled using a standard region filling algorithm.
  This sequence also displays some improvements of the
  adversarial fine-tuning on the car contours. More examples are present in
  the supplementary material, where we can observe that difficult cases for our method include dealing with occlusions and with fast ego-motion.

\begin{figure}
  \begin{center}
\includegraphics[width=\linewidth]{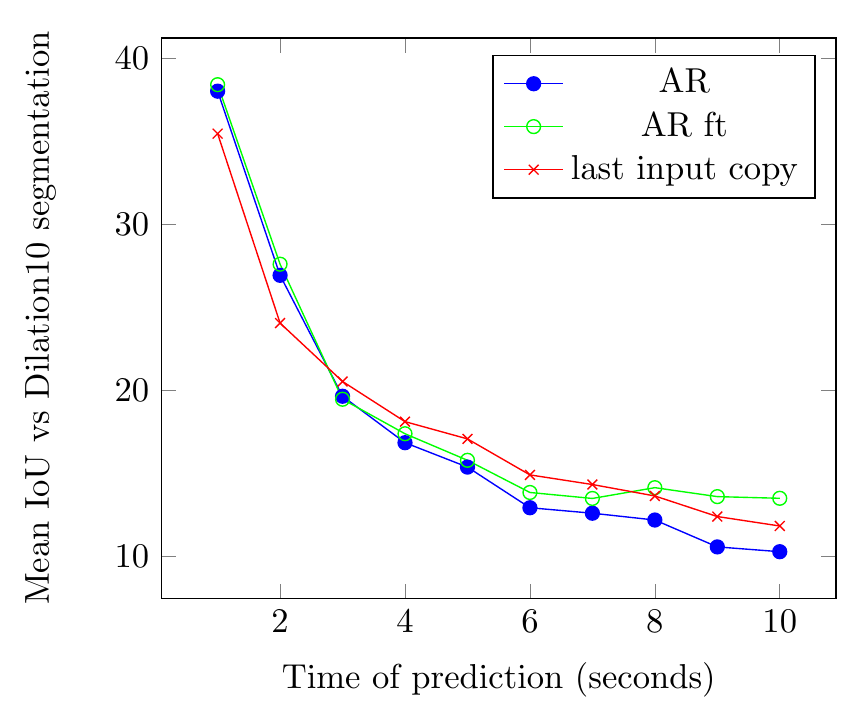}
\end{center}
\myfigcapspace
   \caption{Mean IoU SEG of long-term segmentation prediction for the AR and AR fine-tune \model{S}{S} models.}
\label{curve:long}
\myfigspace
\end{figure}

\subsection{Long-term prediction}
\label{sec:long}

To evaluate the limits of our \model{S}{S} autoregressive models on arbitrarily long sequences, we use them to make predictions of up to ten seconds into the future.
To this end, we evaluate our models on ten sequences on 238 frames extracted from the long Frankfurt sequence of the Cityscapes validation set.
Given four
segmentation frames with a frame interval of 17 images, corresponding to exactly one second, we apply our models to
predict the ten next ones.
  In Figure~\ref{curve:long} we report the IoU SEG
performance as a function of time.  In this extremely challenging setting, the
predictive performance quickly drops over time.
Fine-tuning the model in
autoregressive mode improves its performance,
but only gives a clear advantage
over the input-copy baseline for predictions at one and two seconds ahead.
We also applied our model with a frame interval of 3 to
predict up to 55 steps ahead, but found this to perform much worse.
Figure~\ref{fig:long} shows an example of predictions compared to the actual
future segmentations.
The visualization shows that our model
averages the different classes into an average future, which is perhaps not entirely surprising.
Sampling different possible futures using a GAN or VAE approach might be a way to resolve this issue.

\subsection{Cross-dataset generalization}

To evaluate the generalization capacity of our approach,
we test our \model{S}{S} model on the Camvid
dataset~\cite{Brostow08Camvid}, specifically on the test set of 233 images with 11 classes grouping employed in \cite{badrinarayanan2015segnet}. Ground truth
segmentations are provided for every second on 30 fps video sequences.
We first generate the Dilation10 segmentations - without fine-tuning the oracle to the CamVid dataset - using a frame interval of 5, roughly corresponding to a frame interval of 3 on Cityscapes. We note that the class correspondence between Cityscapes and CamVid is not perfect; for instance we associate the class ``tree'' to ``vegetation''.
As reported in Table~\ref{tab:shorttermcamvidsmall}, our models have very good mid-term performance on this dataset, considering the oracle results.
For reference, \cite{YuKoltun2016} reports an IoU of 65.3 using a fine-tuned Dilation8.

\begin{table}
	{\small
\begin{center}
\begin{tabular}{lcccc}
\toprule
& Dilation10 & Copy last & Warp last& \model{S}{S}\\
& oracle &   input  & input & AR ft \\
\midrule
IoU GT & 55.4 & 40.8  & 43.7 & {\bf 46.8} \\
\bottomrule
\end{tabular}
\end{center}
}
\mytabcapspace
\caption{IoU of oracle and mid-term predictions on Camvid}
   \label{tab:shorttermcamvidsmall}
   \mytabspace
\end{table}

\omitme{
\begin{table}
{\small
\begin{center}
\begin{tabular}{lccc}
\toprule
Model & IoU GT & pp acc. & pc acc. \\
\midrule
Dilation10 oracle & 55.4  & 85.7 & 69.8\\
\midrule
Copy last input short term & 47.3 & 82.4  & 61.2 \\
Warp last input short term & 50.9 & 84.0 & {\bf 65.2} \\
Short term \model{S}{S} & {\bf 51.9}  & {\bf 84.3} & 64.8 \\
\midrule
Copy last input mid-term & 40.8 & 78.4  & 54.3 \\
Warp last input mid-term & 43.7   & 80.1 & 58.2 \\
Mid term \model{S}{S} AR ft & {\bf 46.8} & {\bf 82.6}  & {\bf 58.3}  \\
\bottomrule
\end{tabular}
\end{center}
}
\mytabcapspace
\caption{Segmentation predictions on Camvid sequences.}
   \label{tab:shorttermcamvid}
   \mytabspace
\end{table}
}


\section{Conclusion}
\label{sec:conclusion}

We introduced a new visual understanding task of predicting future semantic
segmentations.  For prediction beyond a single frame, we considered batch models
that predict all future frames at once, and autoregressive models that
sequentially predict the future frames.  While batch models were more effective
in the RGB intensities space because of otherwise large error propagation, the
more desirable autoregressive mode was more accurate in the semantic
segmentation space, supporting with experimental evidence our motivation for
this new task. The autoregressive mode lends itself naturally to predicting
sequences of arbitrary length, thanks to which we can aim to model more
interesting distributions.

In this respect, there is still room for improvement. Where the Dilation10
network for semantic image segmentation gives around 69 IoU, this drops to about
59 when predicting 0.18s ahead and to about 48 for 0.5s. Most predicted object
trajectories are reasonable, but do not always correspond to the actual observed
trajectories.  GAN or VAE models may be useful to address the inherent
uncertainty in the prediction of future segmentations. We open-source our
Torch-based implementation, and invite the reader to watch videos of our
predictions at \url{https://thoth.inrialpes.fr/people/pluc/iccv2017}.

\medskip
{{\bf Acknowledgment.}
This work has been partially supported by the grant ANR-16-CE23-0006 ``Deep in
France'' and LabEx PERSYVAL-Lab (ANR-11-LABX-0025-01). We thank Michael Mathieu,
Matthijs Douze, Herv\'e Jegou, Larry Zitnick, Moustapha Cisse,
Gabriel Synnaeve and anonymous reviewers for their precious comments.
}

{\small
\bibliographystyle{ieee}
\bibliography{egbib,jjv}
}

\end{document}